\pdfoutput=1
\documentclass[11pt]{article}

\usepackage[preprint]{acl}

\usepackage{times}
\usepackage{latexsym}

\usepackage[T1]{fontenc}
\usepackage[utf8]{inputenc}

\usepackage{microtype}

\usepackage{inconsolata}

\usepackage{multirow}

\usepackage{listings}
\usepackage{tikz}
\usepackage{caption}
\usepackage{subcaption}
\usepackage{comment}
\usepackage[noend]{algorithmic}
\usepackage{algorithm}
\usepackage{amsmath,amssymb}
\usepackage{amsfonts}
\usepackage{graphicx}
\usepackage[normalem]{ulem}
\usepackage{enumitem}
\usepackage{amsthm}
\usepackage{bbm}
\usepackage{ulem}
\newtheorem{theorem}{Theorem}[section]
\newtheorem{proposition}[theorem]{Proposition}

\newtheorem{definition}[theorem]{Definition}
\newtheorem{assumption}[theorem]{Assumption}
\newtheorem{example}[theorem]{Example}
\newcommand{\yp}[1]{\textcolor{blue}{Yipeng: #1}}

\newcommand{\red}[1]{\textcolor{red}{#1}}

\newcommand{\D}{\ensuremath{\mathbf{D}}}
\renewcommand{\l}{\ensuremath{\mathtt{l}}}
\renewcommand{\r}{\ensuremath{\mathtt{r}}}
\newcommand{\KB}{\ensuremath{\mathtt{KB}}}
\newcommand{\DOC}{\ensuremath{\mathtt{DOC}}}
\newcommand{\BACK}{\ensuremath{\mathtt{BG}}}
\newcommand{\PC}{\ensuremath{\mathtt{PC}}}
\title{Causal Graph Discovery with Retrieval-Augmented Generation based Large Language Models}

\author{Yuzhe Zhang\thanks{Equal contribution}\thanks{Email address: [first name].[last name]@data61.csiro.au}$^1$ \And Yipeng Zhang$^{* \dagger 1}$ \And Yidong Gan\thanks{Email address: yidong.gan@sydney.edu.au}$^{*1,2}$ \\ \ \\ \textsuperscript{1}Data61, CSIRO\\\textsuperscript{2}University of Sydney\\\textsuperscript{3}University of New South Wales\\
Sydney, Australia\And Lina Yao$^{\dagger 1,3}$ \And Chen Wang$^{\dagger 1}$ }

\begin{document}
\maketitle
\begin{abstract}
%Causal graph construction is essential in the field of causal inference \cite{Pearl2000causal}. 
Causal graph recovery is traditionally done using statistical estimation-based methods or based on individual's knowledge about variables of interests. They often suffer from
%Traditional methods for constructing causal graphs typically rely on either human expertise or data-driven(\red{numerical-data-driven?}\cw{statistical ML-based vs. knowledge-based}) algorithms, \cw{data-driven is too general here (LLM-based method is also data-driven). consider a more specific term?}  
data collection biases and limitations of individuals' knowledge. The advance of large language models (LLMs) provides opportunities to address these problems. %vulnerable to a range of errors and biases due to the limitations inherent in individual research efforts, such as restricted domain knowledge, methodological shortcomings, and various forms of data bias. 
We propose a novel method that leverages LLMs to deduce causal relationships in general causal graph recovery tasks. This method leverages knowledge compressed in LLMs and knowledge LLMs extracted from scientific publication database as well as experiment data about factors of interest to achieve this goal. %leverages Retrieval Augmented-Generation (RAG) based LLMs to systematically analyze and extract pertinent information from a comprehensive collection of research papers. 
Our method gives a prompting strategy to extract associational relationships among those factors and a mechanism to perform causality verification for these associations. %uses a prompt strategy elaborately designed according to the principles of statistical estimation-based methods, such that the LLMs can accurately verify the causal relationships by aggregating their identified associational relationships, 
Comparing to other LLM-based methods that directly instruct LLMs to do the highly complex causal reasoning, our method shows clear advantage on causal graph quality on benchmark datasets. More importantly, as causality among some factors may change as new research results emerge, our method show sensitivity to new evidence in the literature and can provide useful information for updating causal graphs accordingly. % suggestions revising these widely used ground truth based on the solid evidence returned by our method.
%Our method first retrieves relevant text \oran{chunks} from the aggregated literature. Then, the LLM is tasked with identifying and labelling potential associations between factors. %, which is much more straightforward than directly identifying causal relationships. 
%Finally, we give a method to aggregate the associational relationships to build a causal graph. We demonstrate our method is able to construct high-quality causal graphs on the well-known SACHS dataset solely from literature. % based on literaturethe efficacy of our method, we validate it against the well-established SACHS dataset and conduct a comparative analysis with existing data-driven causal discovery algorithms.
%\yz{To rewrite.}
\end{abstract}

\section{Introduction}
%\yp{I have rewritten this part to improve fluency, check if it is appropriate.}

Estimating causal effect between variables from observational data is a fundamental problem to many domains including medical science \cite{hofler2005causal}, social science \cite{angrist1996identification}, and economics \cite{Imbens_Rubin_2015, yao2021survey}. %, and various computational domains\red{cite}. 
It enables reliable decision-making from complex data with entangled associations. %disentangle complex data structures and investigate causality.

While it is usually expensive and infeasible to investigate causal effects by the golden standard---randomized experiments---researchers employ causal inference \cite{pearl2010causal} to estimate causal effects from observational data.
%To determine causal effects from observational data, researchers employ causal inference.
There are two main frameworks for causal inference: the potential outcome framework~\cite{rubin1974estimating} and the structural causal model (SCM)~\cite{pearl1995causal}.
%While both frameworks aim to discover causal relationships, SCMs, and particularly Directed Graphical Causal Models (DGCMs) \cite{Pearl2000causal, spirtes2001}, offer a graphical approach to representing and analyzing these relationships.  
Priori causal structures, usually represented as Directed Graphical Causal Models (DGCMs) \cite{Pearl2000causal, spirtes2001}, are often used to represent and analyze the causal relationships.
%a is a powerful SCM method for representing and analyzing the causal relationships among factors. %These models can be applied to a variety of data types, including cross-sectional, longitudinal, and time-series data. 
%Causal graphs, which are integral to DGCMs, visually depict the hypothesized causal connections between nodes (factors) with directed edges.
%Causal graphs, a core component of DGCMs, effectively illustrate hypothesized causal connections between variables (nodes) using directed edges.
These causal graphs help disentangle the complex interdependencies and facilitate the analysis of causal effects.
%Therefore, recovering
%the structure of 
%causal graphs is essential for understanding the underlying causal mechanisms. 
Recovering causal graphs often relies on experts' knowledge or statistical estimation on experimental data~\cite{doi:10.1177/089443939100900106}.
%Causal graph recovery~\cite{doi:10.1177/089443939100900106} usually seeks information from domain knowledge or data to uncover the structure of causal graphs.
%and is often done through Causal Discovery (CD) \cite{glymour2019review} methods using a statistical estimation-based approach through observational data analysis when interventions or randomized experiments are not viable.
%Causal Discovery (CD) \cite{glymour2019review} methods are integral to DGCMs, providing a data-driven approach to uncover the causal structures from observational data when randomized experiments are not viable. Practically, randomized experiments are often expensive or infeasible, and CD methods offer a more cost-effective alternative.
Causal Discovery (CD) algorithms ~\cite{doi:10.1177/089443939100900106} are the main statistical estimation-based methods that use conditional independence tests to assess associational relationships (called associational reasoning) for inferring causal connections~\cite{spirtes2001, chickering2002optimal, shimizu2006linear, sanchez2018causal}.
%that uncover the structure of causal graphs from observational data.

%methods are employed to uncover the structure of causal graphs from observational data when interventions or randomized experiments are infeasible, which c

%Various algorithms along this line \cite{spirtes2001, chickering2002optimal, shimizu2006linear, sanchez2018causal} utilize statistical tests to assess associational relationships between factors as evidence to infer causal connections.
Consequently, the reliability of these algorithms is affected by the quality of data, which can be compromised by issues such as measurement error \cite{zhang2017causal}, selection bias \cite{Bareinboim2014} and unmeasured confounders~\cite{DBLP:conf/aistats/BhattacharyaNMS21} (See Example~\ref{ex:bias} in Appendix~\ref{sec:ex}).
Additionally, CD algorithms often assume certain distribution, such as Gaussian about data, which may fail to accurately reflect the complexity of real-world scenarios. 
%These shortcomings contribute to the susceptibility of CD methods to biases arising from both the data collection process and the model assumptions, underscoring the need for careful consideration and validation of the methods used in causal inference.

%LLMs, which have seen substantial improvements in their capabilities, are increasingly being employed in methods to predict causal relations \red{cite}.
%These methods leverage LLMs' extensive knowledge base and sophisticated reasoning capabilities to analyze vast amounts of text and infer causal relations.
%However, LLMs are not immune to the biases present in their training data, and their ability to perform causal reasoning is still under scrutiny \cite{zevcevic2023causal} \red{citemore}.
%Moreover, when LLMs are prompted to infer causality from the individual scientific literature, they face the additional challenge of protecting casual relations from potentially biased information \cite{wadhwa2023revisiting}.

To mitigate the limitations, % of data quality in statistical estimation-based CD methods and limited human expert scope in expert-based methods, 
Large Language Models (LLMs) \cite{zhao2023survey} have recently been employed for causal graph recovery~\cite{zhou2023emerging}. There are two main streams of these work:
1) directly outputting causal graphs \cite{choi2022lmpriors, long2022can, kiciman2023causal}; 2) assisting in refining causal graphs generated by statistical estimation-based methods \citet{vashishtha2023causal, ban2023query}.
%Existing LLM-based CD methods mainly utilize LLMs in two ways: recovering causal graphs purely relying on LLMs, or using LLMs to improve statistics/ML-based methods.
%More notable works in these streams include \cite{jiralerspong2024efficient}, that proposes a breadth-first search approach to reduce the complexity of the first stream method, and \citet{vashishtha2023causal, ban2023query} employ LLMs to inject domain causal knowledge into statistical estimation-based outputs, belonging to the second stream.
Most work have a straightforward way of using LLMs. They directly query the causal relationship between each pair of variables \cite{choi2022lmpriors, long2022can, kiciman2023causal} by prompting LLMs with the definition of causality, task details and description of the variables of interest.
%Alternatively, \citet{vashishtha2023causal, ban2023query} employ LLMs to inject domain causal knowledge into statistical estimation-based methods, yet similar issues exist with these methods.
%to recover causal graphs.
They require LLMs to have extensive domain knowledge and capabilities to perform complex causal reasoning. Whether LLMs have sufficient knowledge in specific domains or whether they have causal reasoning capabilities are questionable\cite{kandpal2023large, zevcevic2023causal}.
%However, such methods require LLMs to have extensive background knowledge and robust causal reasoning skills, which hind highly complex inferring process and are still being critically assessed \cite{zevcevic2023causal}.

An alternative approach is to exploit LLMs' capabilities on associational reasoning, e.g., querying the conditional independences (CIs) and recover causal graphs based on extracted associations using CD algorithms\cite{cohrs2023large}. %The causal relationships can be inferred based on extracted associations following CD algorithms.
%Among only few works in this line, \cite{cohrs2023large} queries the LLMs the CIs and construct the causal graph,
However, it remains difficult for LLMs to understand the CIs between variables, especially the independence conditioned on a large set of variables.
\cite{jiralerspong2024efficient} tries to inject statistical CI results into LLMs to improve direct causal relationship query results, but the efficacy varies among datasets.
%To solve the high complexity issue, \citet{jiralerspong2024efficient} proposed a breadth-first search approach to reduce the number of required queries.
%Alternatively, \citet{vashishtha2023causal, ban2023query} employ LLMs to inject domain causal knowledge into statistical estimation-based methods, yet similar issues exist with these methods. %such methods suffer from the same issues.

%\oran{To mitigate the limitations of data quality in statistical estimation-based causal graph recovery tasks, we propose to \red{recover causal connections by information extracted}
%extract causal relations 
%from a knowledge base containing related literature, which contains valuable insight hidden in datasets about \red{associational/causal} relationships among variables. % of which is a representation of information hidden in a set of data.
%The elaborately aggregated information can potentially overcome shortcomings of statistical estimation-based approaches.
%We leverage Large Language Models (LLMs) \cite{zhao2023survey} to accomplish the information extraction from large document databases.}

We propose the LLM Assisted Causal Recovery (LACR) method to address these challenges. LACR enhances the knowledge base of LLMs with Retrieval Augmented Generation (RAG) \cite{lewis2020retrieval, borgeaud2022improving} for reliable associational reasoning. We retrieve highly related knowledge base from a large scientific corpus that contains valuable insight hidden in datasets about associational/causal relationships among variables.
We further enhance the accuracy of LCAR's causal recovery results by aggregating the collective extracted information from related literature according the Wisdom of the Crowd principle \cite{Grofman1983ThirteenTI}.
%propose the LLM Assisted Causal Recovery (LACR) method to recover causal graphs by leveraging Retrieval Augmented Generation (RAG) \cite{lewis2020retrieval, borgeaud2022improving} for systematic associational reasoning based on reliable retrieved knowledge base, and inferring causal relationships with CD algorithms.
LACR also uses an associaitonal reasoning-based causal recovery prompt strategy which elaborately instructs the LLMs the mathematical intuitions behind conditional independence, and builds a surjection from conditional independences extracted by LLMs to causal relationships between variables. 
%To ensure the accuracy and reliability of LLMs' associational reasoning, we retrieve highly related knowledge base from a large scientific corpus that contains valuable insight hidden in datasets about associational/causal relationships among variables.
%We further enhance the accuracy of LCAR's causal recovery results by aggregating the collective extracted information from related literature according the Wisdom of the Crowd principle \cite{Grofman1983ThirteenTI}.
LACR is data-driven and dos not rely on task-specific knowledge for document retrieval or prompt design. It can serve as a causal graph recovery tool for generic tasks.

Our methodology provides a structured and systematic approach to inferring causal relationships, as it is grounded in a broader evidentiary base and subject to systematic validation.
As LACR conducts associational reasoning on a reliable knowledge base, most of which provide evidences based on experimental data analysis, LACR largely overcomes the collection bias problem in statistical estimation-based CD algorithms.
We discuss this in detail in Section~\ref{sec:experiment} by pointing out the causal conflict between the well-known causal discovery results and recent research results extracted by LACR.

\vspace{.5em}
%\subsection{Contributions}
\noindent\textbf{Our Contributions:}
%\medskip
%\begin{enumerate}[leftmargin=*, itemsep=0pt, parsep=0pt, labelwidth=1.25em, itemindent=1.25em]
%\noindent 

\noindent\textit{$\bullet$} We introduce a novel RAG-based causal graph recovery method that achieves better associational reasoning. The method shows its potential in accurate causal graph construction and overcoming data collection bias issues in traditional methods.

\noindent\textit{$\bullet$} We design an associational reasoning-based prompting strategy that reduce LLMs' task complexity to simple associational reasoning to improve the reliability of LLMs' results. The reliability gain further improve the quality of recovered causasl graphs. %hich proves good performance without manual adjusting.

\noindent\textit{$\bullet$} We conduct experiments in several well-known real-world causal graphs and demonstrate the efficacy of LACR. More importantly, based on the scientific evidence returned by our method, we show bias exists in the validation datasets widely used in the CD community, and suggest ways to improve.

\section{Background}\label{sec:preliminary}

In this section, we introduce the preliminaries of the {\em directed graphical causal models} (DGCM) and the {\em causal graph recovery} problem.

\subsection{Directed Graphical Causal Models}

A {\em Directed Graphical Causal Model} (DGCM) is a tuple $M=\langle G, P \rangle$. In the model, $G=\langle V, E \rangle$ is a Directed Acyclic Graph (DAG), also known as a {\em causal graph}, where the set of nodes $V=\{v_1, \cdots, v_n\}$ represents random variables (with $|V|=n$), and $E\subseteq \{(v_i,v_j)\mid v_i,v_j\in V, v_i\not= v_j\}$ is a set of directed edges, also called {\em causal edges}, that encode {\em causal relationships}. 
Let $\bar{G}=\langle V, \bar{E}\rangle$ be the {\em skeleton} of DAG $G$, where each $(v_i, v_j)\in \bar{E}$ is an undirected edge, and it indicates that one of $(v_i, v_j)$ and $(v_j, v_i)$ is in $E$. 
Let a sequence of distinct nodes $\ell = (v_{j_1}, v_{j_2}, \cdots, v_{j_m})$ denote a {\em path}, such that for each $i \in \{1,2,\cdots, m-1\}$, $(v_{j_i},v_{j_{i+1}}) \in \bar{E}$.
A path is a {\em causal path} from $v_{j_1}$ to $v_{j_{m}}$ if for each $i \in \{1,2,\cdots, m-1\}$, $(v_{j_i},v_{j_{i+1}}) \in E$.
The joint probability distribution of all variables is denoted by $P$.
%A directed path $\ell$ is a sequence of nodes $(v_{i_1}, v_{i_2}, \cdots, v_{i_k})$, where each $v_{i_j}$ is a node in $V$ and for each $j \in {1,2,\cdots, k-1}$, $(v_{i_j}, v_{i_{j+1}}) \in E$. Here, the subscript $i_j$ is an index that identifies the position of the node within the path, and $k$ denotes the length of the path.
Note that we do not consider any variable other than those in $V$, that is, we assume there is no so-called latent or exogenous variable.
\noindent\textbf{Constraints of causal graphs.}
A causal graph is subject to a series of constraints on variables' {\em associational relationships}.
Especially, the causal edges specify the causal relationships between variables.
Given $(v_i,v_j)\in E$, $v_i$ is a {\it direct cause} of $v_j$.
That is, when holding the other variables constant, varying the value of $v_i$ triggers a corresponding change in the value of $v_j$, but not vice versa.
This causal relationship thus entails the associational relationship between the variables, i.e., their marginal probability distributions $P(v_i)$ and $P(v_j)$ are associated (or correlated), which does not have the direction attribute.
Notice that two variables can be associated even though they do not have a direct causal relationship between each other.
Typical examples are that two variables linked by a causal path, and two variables pointed to by two causal paths that have the same starting node (which is usually called a covariate).
%the two variables have an indirect causal relationship through other variables, or they share the same parent node in $G$, which is usually called {\em confounding} in causal inference.
%Consequently, the marginal probability distributions $P(v_i)$ and $P(v_j)$ are correlated.
The precise constraints follow an assumption of the causal graph called the Causal Markov Assumption.
\begin{assumption}[Causal Markov Assumption]\label{asm:markov}
In any causal graph, each variable is independent of its non-descendants conditioned on its parents in the causal graph.
% it holds that
% \begin{align}
%     %P_{v_1,v_2,\cdots, v_n}(v_1,\cdots, v_n)=\prod_{v_i\in V}P_{v_i\mid pa_i}(v_i\mid pa_i),
%     P(v_1, v_2, \cdots, v_n) = \prod_{v_i \in V} P(v_i \mid pa(v_i)),
% \end{align}
% where $pa(v_i)$ is the set of parents of $v_i$ in the causal graph.
\end{assumption}

Therefore, the structure of a causal graph implies graphical constraints called {\em d-separation} \cite{Pearl2000causal} that specify a conditional associational relationship between variables.
In the rest of this paper, for any given variable pair $v_i, v_j\in V$, we constantly use $V'$ to denote an arbitrary subset of $V\setminus \{v_i, v_j\}$, unless otherwise specified.
\begin{definition}[d-separation]\label{def:dsep}
A variable set $V'$ {\it blocks} a path $\ell$ if (i) $\ell$ contains at least one arrow-emitting variable belonging to $V'$, or (ii) $\ell$ contains at least one collider (variable $v_i$ is a collider if $(v_{j_{i-1}}, v_{j_i}), (v_{j_{i+1}}, v_{j_i})\in E$)
%\sout{, where $v_{j_{i-1}}, v_{j_i}$, and $v_{j_{i+1}}$ are three adjacent nodes on $\ell$}\yp{$v_{j_{i-1}}, v_{j_i}$, and $v_{j_{i+1}}$ are not three adjacent nodes, and we already define that $\ell$ is a sequence of nodes.}) 
that does not belong to $V'$ and has no descendant belonging to $V'$.
If $V'$ blocks all paths from $v_i$ to $v_j$, $V'$ is said to {\it d-separate} $v_i$ and $v_j$.
%\yp{since $\ell$ can include edges in both directions between two nodes, what is the meaning of the arrow-emitting variable and collider? Should we use the causal path instead of the path?}
%denoted as $v_i \indep v_j\mid S$.
\end{definition}
If $V'$ d-separates $v_i$ and $v_j$, then the joint probability distribution $P$ encodes that the two variables are independent conditioned on $V'$.%, denoted as $v_i\perp v_j\mid V'$.

% \paragraph{Assumptions of causal graphs}
% The Markov property of the causal graph interprets that d-separation in causal graphs indicates conditional independence between variables.% can be interpreted as the Markov property of the causal graph.
% This is a necessary assumption based on which the DGCM works.
% \begin{assumption}[Causal Markov Assumption]\label{asm:markov}
% In each DGCM, each variable is independent of its non-descendants conditioned on its parents in the causal graph.
% it holds that
% \begin{align}
%     %P_{v_1,v_2,\cdots, v_n}(v_1,\cdots, v_n)=\prod_{v_i\in V}P_{v_i\mid pa_i}(v_i\mid pa_i),
%     P(v_1, v_2, \cdots, v_n) = \prod_{v_i \in V} P(v_i \mid pa(v_i)),
% \end{align}
% where $pa(v_i)$ is the set of parents of $v_i$ in the causal graph.
% \end{assumption}
% Intuitively, in a DGCM, each variable is independent of its non-descendants conditioned on its parents in the causal graph.

Assumption~\ref{asm:markov} is a necessary condition for the encoding of the associaitonal relationship constraints in $P$.
On the other hand, the following {\em faithfulness assumption} is a sufficient condition that $P$ encodes such constraints.
% Practically, the joint probability distribution may contain additional independent information that is not induced by the d-separation constraints.
% Due to the sake of DGCM's validation, we assume that there is no such additional independency information, formalized as the following assumption.
\begin{assumption}[Causal Faithfulness Assumption]\label{asm:faith}
A joint distribution $P$ does not encode additional conditional associational relationships other than those consistent with $G$'s d-separation information.
We call such $P$ is faithful to $G$.
%or dataset $\D$ generated from a causal graph $G$, do not encode additional conditional independent relationships other than those consistent to $G$'s d-separation information.
%We call such $P_V$ and $\D$ are faithful to $G$.
\end{assumption}

We now formally define the constraints that follow distribution $P$ faithful to causal graph $G$. 
%Given a DGCM $M=\langle G, P \rangle$, with Assumptions \ref{asm:markov} and \ref{asm:faith}, we have that variables' marginal probabilities follow the conditional associational relationship indicated by $G$.
Let $\alpha(ij\mid V')\in \{0,1\}$ be the {\em conditional associational relationship} between variables $v_i, v_j\in V$ conditioned on variable set $V'$.
$\alpha(ij\mid V')=0$ denotes that $v_i$ and $v_j$ are independent conditioned on $V'$ according to $P$, and $\alpha(ij\mid V')=1$ denotes associated.
%For the remainder of this paper, unless otherwise specified, $V'$  will refer to any subset of $V \setminus \{v_i, v_j\}$. \yp{check}%\yp{Is V' equal to S in d-separation? Should we keep it consistent?} \yz{Changed}
We write $\alpha(ij)$ when $V'=\emptyset$.

Then, by Assumptions \ref{asm:markov} and \ref{asm:faith} and Definition~\ref{def:dsep}, we have that for $v_i, v_j\in V$:

\noindent 1. $V'$ d-separates $v_i$ and $v_j \implies  \alpha(ij\mid V')=0$;\\
2. $\alpha(ij)=1$ and $(v_i, v_j)\notin \bar{E} \implies \exists V'$ s.t. $ \alpha(ij\mid V')=0$;\\
3. $(v_i, v_j)\in \bar{E} \implies \nexists V'$ s.t. $\alpha(ij\mid V') = 0$.%, there does not exist $V'\subseteq V\setminus \{v_i, v_j\}$ such that $\alpha_{ij\mid V'}=0$.

\section{Methodology}\label{sec:method}
\begin{comment}
\begin{itemize}
\item Existing LLM methods query $\zeta_{ij}$ and $\delta_{ij}$, which are not trivial to obtain. (\red{Why?})
\item We query a document (which contains $\alpha_{ij\mid V'}(\D')$) to give a noisy $\delta_{ij}(\alpha_{ij}(\D))=\r$.
\item Aggregating $\alpha_{ij\mid V'}(\D')$, $\alpha_{ij\mid V'}(\D'')$ ... can mitigate bias (theorem).
\item What do we extract? Do we have $\alpha_{ij\mid V'}$ for all $V'\subseteq V\setminus \{v_i, v_j\}$? If not, how do we deal with it? E.g., assume indirect association indicating conditional independence?
\item Mapping.
\end{itemize}
\end{comment}
We now start to introduce our LLM-based method, called {\em large language model assisted causal recovery} (LACR), that uses a prompt strategy elaborately designed following the process of a statistical estimation-based CD method, called the {\em constraint-based causal graph construction} (CCGC).
We first show how CCGC works.

\subsection{Constraint-based Causal Graph Construction: From Data to Causation}\label{sec:constraint_based_CD}
Based on Assumptions \ref{asm:markov} and \ref{asm:faith}, we are able to partially construct the causal graph $G$ from a {\em knowledge base} $\KB$ that is faithful to $G$ by a statistical estimation-based method.
In a nutshell, $\KB$ can be but not limited to data, the LLM's background knowledge, and external documents.
For more details, see Section~\ref{sec:ccc}.
We take data as the $\KB$ in CCGC.
A $\KB$ is called faithful to $G$ if it estimates a joint distribution that is faithful to $G$.

The process of CCGC can be divided into two phases: the {\em edge existence verification} phase, which first constructs the skeleton, and the {\em orientation} phase, which determines the direction of each undirected edge.
LACR only uses the CCGC-based prompt strategy to conduct the edge existence verification, and therefore, we only introduce the first phase of CCGC.
%In our method, we rely on the LLMs to determine the associational relationships between variables to construct the skeleton based on the process of CCGC, and then, directly query LLMs the direction of each edge in the skeleton.
%Therefore, we introduce the first phase of CCGC, namely, the edge existence verification.

%\red{cite recover Markov equivalent class paper} specifies that a DAG may not be recovered even with sufficient data, and instead, we can only guarantee to recover a Markov equivalent class, i.e., a set of DAGs such that each of them corresponds to the same conditional independence relationships between variables.
%In other words, a sufficient dataset may not contain enough information to determine the direction of each edge, even though we can infer the existence of each undirected edge.
%Therefore, we formulate the causal discovery process in two phases: the {\em edge existence verification} phase and the {\em orientation} phase.\\
%\red{Mention model assumption? E.g., linear or non-parametric.}\\
%\red{Notation: association $\alpha$, causation $\zeta$, direction $\delta$.}

%{\bf Edge existence verification}
For each pair of variables $v_i, v_j\in V$, we verify the existence of the undirected edge in between (i.e., whether $(v_i, v_j)\in \bar{E}$ or not) by statistically testing whether $v_i$ and $v_j$ can be d-separated by any variable set $V'$. %any subset of $V\setminus \{v_i, v_j\}$.
%Let $\alphaso : (\mathcal{D}\rightarrow V\times V)\rightarrow \{0,1\}^{2^{n-2}}$ be an estimator of conditional associational relationship between variable pairs, where $\mathcal{D}$ is the space of $n$-dimensional datasets which are faithful to a causal graph $G$.
Let $\hat{\alpha}_{\KB}(ij\mid V') \in \{0,1\}$ be an estimator of $\alpha(ij\mid V')$, based on $\KB$.
%That is, $\alpha_{ij\mid V'} (\D)=0$ denotes that $v_i$ and $v_j$ are tested independent from each other conditioned on variables in $V'$ based on dataset $\D$, and $\alpha_{ij\mid V'} (\D)=1$ for not independent.
Next, based on a given $\KB$ that is faithful to $G$, we define $\zeta_{\KB}:V\times V\rightarrow \{0,1\}$ as the {\em causal edge existence} mapping, such that $\zeta_{\KB}(ij)=0$ if $\exists V'\text{ s.t. }\ \hat{\alpha}_{\KB}(ij\mid V')=0$, otherwise $\zeta_{\KB}(ij)=1$.
% \begin{align}\label{eq:verify}
% \zeta_{\KB}(ij)=\begin{cases}
% 0,\ \text{if } \exists V'\text{ s.t. }\ \hat{\alpha}_{\KB}(ij\mid V')=0\\
% 1,\ \text{otherwise} \end{cases}.
% \end{align}
%where $V'\subseteq V\setminus\{v_i, v_j\}$ and $\mathcal{D}$ 
%where $\D$ is an arbitrary dataset that is faithful to $G$. 
$\zeta_{\KB}(ij) = 0$ implies that we estimate there is no edge between $v_i$ and $v_j$, since the pair of variables can be d-separated by at least one variable set.
See Appendix \ref{sec:ex} for an example of CCGC's process.

Compared with LACR, most existing LLM-based causal graph construction methods directly query LLMs the causal relationships.
With the introduction of CCGC, we next illustrate the limited reliability of such methods.\\

\noindent{\bf Limited Reliability of Direct Causal Prompt.}
We name the prompt used in such direct query of causal relationships as the {\em direct causal prompt}.
Examples include ``Is A a cause of B?'' and ``Does the change of A cause the change of B'', which are wildly used in related work~\cite{kiciman2023causal, choi2022lmpriors, long2022can}.
Such prompt directly queries the causal edge existence ($\zeta_{\KB}(\cdot)$) and the causal direction.% ($\delta_{\KB}(\cdot)$).
We argue that such direct prompting requires extensive causal reasoning capability from LLMs.
%There is evidence showing LLMs are not causal~\cite{zevcevic2023causal}. However, they can be used as a stepping stone to infer causal relationships. %, as the inference of such causal relationships is computational complex.
The following proposition shows the high complexity hidden behind a direct causal prompt.
\begin{proposition}\label{prop:complex}
Assuming that estimating $\hat{\alpha}_{\KB}(ij\mid V')$ for a given $V'$ needs $O(1)$ time, inferring $\zeta_{\KB}(ij)$ requires $O(2^{n-2})$, where $n=|V|$.
\end{proposition}

\begin{proof}
To verify if $\zeta_{\KB}(ij)=0$, by the definition of $\zeta(ij)$, we need to check whether there exists a variable set $V'$ such that $V'$ d-separates $v_i$ and $v_j$.
That is, $\alpha_{\KB}(ij\mid V')=0$.
Then, the worst case is that we need to check every combination of $V'\subseteq V\setminus\{v_i, v_j\}$, which needs $O(2^{n-2})$ time.
\end{proof}

We now start formally introducing the {\em large language model assisted causal recovery} (LACR) method, which first extracts the conditional associational relationships between variables, and determines the causal relationships following the process of CCGC (see Section~\ref{sec:constraint_based_CD}).
We implement such a process by a series of separated queries using the constraint-based causal prompt.
%uses LLMs to extract the conditional associational relationships between variables, and then, infer the existence and direction of causal relationships.
The LACR consists of two steps: the edge existence verification (LACR 1) and orientation (LACR 2).
See all of the original prompts in Appendix~\ref{sec:prompt}.

\subsection{LACR 1: Edge Existence Verification}
In this phase, we construct the skeleton of the causal graph, i.e., verifying the existence of each edge without clarifying its direction.
We use LLMs to mine the statistical evidence to verify the conditional associational relationship between each pair of variables and determine the existence of a causal edge (recall Section~\ref{sec:constraint_based_CD}), from the retrieved scientific documents (corresponding to document-based query), LLMs' internal knowledge (corresponding to background-based query), and statistical estimation-based output.
To achieve this target, we design a prompt strategy that encodes the statistical principles of CCGC.

\subsubsection{Constraint-Based Causal (CC) Prompt}~\label{sec:ccc}
%For each variable pair, namely $v_i$ and $v_j$, we clarify their conditional association type by mining the statistical evidence from a series of knowledge bases.
%For each piece of the knowledge base, $\KB$, we conduct a chain of 4 queries to determine the final opinion of the knowledge piece, namely, the {\bf background reminder}, the {\bf association verifier}, the {\bf association type verifier}, and the {\bf association rechecker}.
%See the original prompt in \yz{Appendix}.
%
%Note that $\KB$ is our resource of data in LLM-assisted methods, and therefore, we estimate the conditional associational relationships through it, i.e., $\hat{\alpha}_{ij\mid V'}^\KB$. %\yp{Should we use $\D_\KB$? We may provide different knowledge bases as datasets, and all of them are datasets for the estimator $\hat{\alpha}$}
In LACR, $\KB$ can be the LLM's background knowledge, external documents, and datasets.
%and detailed instructions guiding the LLM in evaluating the association between two variables.
For each variable pair, namely $v_i$ and $v_j$, we clarify their conditional associational relationship by mining the statistical evidence from $\KB$.
We conduct a chain of 4 queries
%(\yp{queries or components?}\yz{Let's keep it queries}) 
to determine the final opinion of each piece of $\KB$, namely, the {\bf background reminder}, the {\bf association verifier}, the {\bf association type verifier}, and the {\bf association rechecker}.
However, if any $\KB$
%the retrieved documents or LLM's background knowledge 
does not contain sufficient information to determine the value of $\hat{\alpha}_{\KB}(ij\mid V')$, we ask the LLM to give an answer {\sc unknown}. We classify such knowledge bases as {\em unusable}, and they are discarded during the decision-making phase.

\vspace{.5em}
\noindent{\bf Background reminder.}
This prompt component helps the LLM to understand the full picture of the task, and avoid misinterpretation of variables' meaning.
We aim to provide minimum external information about the task other than the names of the variables to the LLM.
%Therefore, we only specify the role of the LLM, i.e., designating its role as a {\sc research scientist who tries to find the associational relationships between variable pairs}.
Therefore, we only give the full {\sc factor} list (i.e., the names of the variables), and specify the {\sc domains} from which the variables are.
For example, in the ASIA experiment dataset (see Section~\ref{sec:experiment}), all variables are from the domains of {\sc medical, biology, and social science}.
Finally, we ask the LLM to specify the meaning of each variable, as well as the interaction among them.

\vspace{.5em}
\noindent{\bf Association verifier.}
%This component mines information from $\mathtt{KR}$ to verify the zero order associaitional relationship between $v_i$ and $v_j$, i.e., $\hat{\alpha}_{ij\mid \emptyset}^\KB$.% and simplified to $\hat{\alpha}_{ij}^\KB$ in subsequent discussions.
%In this component, the LLM is asked to thoroughly read the $\KB$, which is either the LLM's background knowledge or a given document, and an instruction on how to judge two factors that are associated, independent, or unknown.
%The principle of the instruction is that, $v_i$ and $v_j$ are associated (resp. independent) if the $\KB$ provides statistical evidence to support or evidence able to statistically support that $\hat{\alpha}_{ij\mid \emptyset}^\KB=1$, i.e., associated (resp. $\hat{\alpha}_{ij\mid \emptyset}^\KB=0$, i.e., independent), otherwise the decision is unknown.
%If the decision is an association, the LACR goes to the next component.
The component utilizes $\mathtt{KR}$ to verify the zero-order associational relationship between $v_i$ and $v_j$, i.e., $\hat{\alpha}_{\KB}(ij)$.
%and referred to as $\hat{\alpha}_{ij\mid \emptyset}^\KB$ in subsequent discussions. 
The LLM is provided with an {\sc association context} (an instruction of how to determine whether $v_i$ and $v_j$ is associated or not) and $\KB$.
Then, the LLM determines the relationship as {\sc associated} (if $\hat{\alpha}_\KB(ij)=1$), {\sc independent} (if $\hat{\alpha}_\KB(ij)=0$), or {\sc unknown}, based on the statistical evidence extracted from $\KB$.
If the decision is an association, the LACR goes to the next query.\\
%\yp{I have rewritten this part, see if it's more readable. There is still a problem, our definition is "$\hat{\alpha}_{ij\mid V'} (\D) \in \{0,1\}$", should we define the "unknown" state?}
%\yz{Guess it's fine without "Unknown". If testing by data, the outcome is binary (independent or not). If the outcome is "Unknown" by LLM, then this $\KB$ is unusable. I wrote this Section 3.2.3. Maybe we can move it around here.}

\noindent{\bf Association type verifier.}
Upon determining {\sc associated} between $v_i$ and $v_j$, we further need to determine whether this association is ``indirect'' or ``direct'', i.e., whether there exists $V'$ that can d-separate $v_i$ and $v_j$.
Based on the given $\KB$ and reasoning, the LLM is asked to read an {\sc accusation type context} (an instruction of how to judge whether the association is indirect or direct based on $\KB$).
Intuitively, the {\sc accusation type context} illustrates that if the association between $v_i$ and $v_j$ is mediated by variables from $V'$ , then, the association is indirect, otherwise it is direct.
To align precisely with CCGC, the {\sc accusation type context} further explains that ``the association mediated by third variables'' means that the association is eliminated if we control the third variables constantly.
The LACR goes to the final query if the decision is an {\sc indirectly associated}.

\vspace{.5em}
\noindent{\bf Association rechecker.}
%Based on the fact that the LLM can possibly determine the association to be indirect because the association is mediated through external variables, i.e., variables not contained in $V$.
%We ask the LLM to check whether the mediating variable set contains those from $V\setminus \{v_i, v_j\}$.
Considering the potential that the LLM can return {\sc indirectly associated} because it judges that the association between $v_i$ and $v_j$ is mediated by external variables that are not from $V$.
Since we do not consider external variables, we ask the LLM to verify whether the set of mediating variables includes any from $V\setminus \{v_i, v_j\}$.
If yes, the association type should be corrected to {\sc directly associated}.
%since we do not consider external factors. 
%, which could lead to an incorrect inference, we ask the LLM to verify whether the set of mediating variables includes any from $V'$. If yes, the association type should be corrected to direct since we do not consider external factors. \yp{check}
%If all variables are external, the association type should be changed to direct since we assume that we do not consider other variables than $V$.
\subsubsection{The CC Prompt is Deterministic}
Using the above prompt strategy, we demonstrate that the LLM's return can determine the existence of causal edges based on a given $\KB$.
%given a $\KB$ (i.e., either the LLM's background knowledge or a given document), we show that each edge's existence can be determined by the LLM's return.
We first specify that the LLM's return based on the above prompt must be one from set \{{\sc independent}, {\sc directly associated}, {\sc indirectly associated}, {\sc unknown}\}.
%yp{using the above prompt, we only know that if $\hat{\alpha}_{ij\mid V'}^\KB = 1$, they are associated, but we are not sure if it is 'Directly' or 'Indirectly'.}
If the return is {\sc Unknown}, the $\KB$ is unusable.
Then, for each {\em usable} $\KB$, we have the following proposition showing that each usable $\KB$ can make decision deterministically.
\begin{proposition}\label{prop:surjection}
For each variable pair $v_i$ and $v_j$, the mapping from the conditional associational relationship space of $v_i$ and $v_j$ to the return set of each usable $\KB$ is a surjection, and the mapping from the return set of each usable $\KB$ to the range of $\zeta_{ij}$, i.e., $\{0,1\}$, is also a surjection.
\end{proposition}

\begin{proof}
For each usable knowledge base $\KB$, any possible return through the CC prompt must from the set \{{\sc directly associated}, {\sc indirectly associated}, {\sc independent}\}.

We first show the first half of the proposition, i.e., the mapping from the conditional associational relationship space between $v_i$ and $v_j$, i.e., $(\hat{\alpha}_\KB(ij\mid V'))_{V' \subseteq V \setminus \{v_i, v_j\}}$, to LLM's return space based on each usable $\KB$, i.e., \{{\sc independent}, {\sc directly associated}, {\sc indirectly associated}\} is a surjection.
Note that $(\hat{\alpha}_\KB(ij\mid V'))_{V' \subseteq V \setminus \{v_i, v_j\}}$ forms a $2^{|V|-2}$-dimensional vector, recording $\alpha_\KB(ij\mid V')\in \{0,1\}$ for all possible $V'$.
We discuss three exclusive cases:
\begin{enumerate}
\item $\hat{\alpha}_\KB(ij)=0$.
That is, the zero-order conditional associational relationship between $v_i$ and $v_j$ is independent.
%It includes that for each $V'\subseteq V\setminus \{v_i, v_j\}$, $\hat{\alpha}_{ij\mid V'}^\KB=0$ by Assumption~\ref{asm:markov}.
This case is mapped to LLM return {\sc independent}.% \yz{Maybe need revising.}
\item For all possible $V'$, $\hat{\alpha}_\KB(ij\mid V')=1$.
The case denotes that $v_i$ and $v_j$ are always associated conditioned on any possible $V'$, and therefore, this case is mapped to LLM return {\sc directly associated}.
\item $\hat{\alpha}_\KB(ij)=1$ and $\exists V'$ such that $|V'|\ge 1$ and $\hat{\alpha}_\KB(ij\mid V')=0$. %\yp{do you mean: $\alpha_{ij\mid V'}=1$ otherwise?}
In this case, controlling variables in $V'$, the statistical association between $v_i$ and $v_j$ is eliminated, and then it is mapped to LLM return {\sc indirect associated}.
\end{enumerate}

We then show the second half of the proposition, i.e., the mapping from \{{\sc independent}, {\sc directly associated}, {\sc indirectly associated}\} to $\{\zeta_\KB(ij)=0, \zeta_\KB(ij)=1\}$ is a surjection.
If the return is {\sc directly associated}, the $\KB$ specifies that $v_i$ and $v_j$ cannot be d-separated, and therefore, it is mapped to $\zeta_{\KB}(ij)=1$.
On the other hand, if LLM return is {\sc independent} or {\sc indirectly associated}, then, it indicates that $V'$ exists that can d-separate $v_i$ and $v_j$, where {\sc independent} corresponds to $V'=\emptyset$.
%there is no causal path between $v_i$ and $v_j$ ({\sc Independent}), or all paths between $v_i$ and $v_j$ can be blocked ({\sc Indirectly Associated}), and
Therefore, these two last cases correspond to $\zeta_{\KB}(ij)=0$.
\end{proof}

\subsubsection{LACR 1}
With the above CC prompt, we are ready to introduce LACR 1 (Algorithm~\ref{algo:LACR 1}).
We initialize the algorithm by setting the skeleton graph $\bar{G}$ as a complete undirected graph $\bar{G}^c$, giving each variable pair $v_i, v_j$ a pre-retrieved set of $k$ relevant scientific documents as the document-based knowledge base $\mathbf{DOC}=\{\mathbf{DOC}_{ij}=\{\DOC_{ij}^1, \cdots, \DOC_{ij}^k\}\}_{v_i,v_j\in V, \text{ s.t., }v_i\not=v_j}$.
Then, for each variable pair, we query the LLM by the CC prompt to estimate $\hat{\zeta}_{\KB}(ij)$ based on each of the given documents provided in $\mathbf{DOC}_{ij}$ and the LLM's background knowledge $\BACK$.
If the decision of the LLM is {\sc indirectly associated} or {\sc independent}, i.e., $\hat{\zeta}_\KB(ij)=0$, by Proposition~\ref{prop:surjection}, we add -1 point to the score $S$, if the decision is {\sc directly associated} (i.e., $\hat{\zeta}_\KB(ij)=1$), we add $1$ point to $S$, otherwise, we do not change $S$ if LLM answers {\sc unknown} based on $\KB$. % (denoted as $\hat{\zeta}_{ij}^{\KB}=-1$). (\yp {subtract 1 point or 0?}\yz{0}\yp{how about remove it?})
After considering all of the LLM's decisions for $v_i$ and $v_j$, if the final score $S>0$, we keep the undirected edge $(v_i, v_j)$; otherwise, we remove it from $\bar{G}$.
Finally, the algorithm returns the skeleton after querying each variable pair based on all $\KB$s.
\begin{algorithm}[t]
\begin{algorithmic}[1]
% \begin{description}
% \item[Input:] $\bar{G} \leftarrow \bar{G}^c$, $\mathbf{DOC}$.
\STATE{{\bf Input:} $\bar{G} \leftarrow \bar{G}^c$, $\mathbf{DOC}$, $\D$}
% \item[Iteration:]\hfill
%\item[Iteration:]\hfill
%\begin{algorithmic}[1]
\FOR{$\forall v_i, v_j\in V$, \text{ s.t., } $v_i\not= v_j$}
\STATE{$S=0$}
%\FOR{$\DOC_{ij}^k \in \mathbf{DOC}_{ij}$}
\FOR{$\KB \in \mathbf{DOC}_{ij}\cup \{\BACK\}$}
% \STATE{$\KB \leftarrow \DOC_{ij}^k \cup\{\BACK\}$}
\IF{$\hat{\zeta}_\KB(ij)=1$}
\STATE{$S += 1$}
\ELSIF {$\hat{\zeta}_\KB(ij)=0$}
\STATE{$S += -1$}
\ENDIF
\ENDFOR
\IF{$S\le 0$}
\STATE{$\bar{G} \leftarrow \bar{G} \backslash (v_i, v_j)$}
\ENDIF
\ENDFOR
\STATE{{\bf Return:} $\bar{G}$}
\end{algorithmic}
% \item[Return:] $\bar{G}$
%\item[Return:] $\bar{G}$
% \end{description}
\caption{LACR 1}
\label{algo:LACR 1}
\end{algorithm}

Note that using the score $S$ to aggregate each $\KB$'s ``opinion'' for each variable pair is equivalent to making the collective decision of $\hat{\zeta}(ij)$ by the simple majority voting rule \cite{brandt2016handbook}.
We slightly bias the decision towards $\hat{\zeta}_\KB(ij)=0$ by the setting of removing an edge if $S\le 0$, since generally, the LLM's decision biases towards {\sc directly associated}.
We use this biased setting because (1) almost $\KB$s cannot load the evidence showing $\hat{\alpha}_\KB(ij\mid V')$ for all possible $V'$, and (2) if most retrieved documents are unusable (no research report on $v_i$ and $v_j$'s association), then, it is more possible that $v_i$ and $v_j$ are not associated.
By the theory of the Wisdom of the Crowd, LACR's decision tends to be more accurate than querying a single knowledge base, and it can be improved by adding more relevant documents (see a detailed description in Appendix~\ref{sec:enhance}).

\paragraph{LACR Expects To Enhance Skeleton Estimation Accuracy}\label{sec:enhance}
The theory of Wisdom of the Crowd \cite{Grofman1983ThirteenTI} states that if (1) each individual voter can make the correct decision better than random decision (e.g., by a toss), and (2) voters make their decision independently, then, the accuracy of the collective decision made by simple majority monotonically increases with the number of voters.
In LACR, each $\KB$ can be seen as a voter.
Generally the above conditions tend to be guaranteed because (1) both $\BACK$ and $\DOC$ have high quality and the delivered information is better than random information, and (2) different research papers deliver their results in a relatively independent way because of scientific integrity.
Therefore, LACR's decision tends to be more accurate than querying single knowledge base, and it can be improved by adding more relevant documents.
%\paragraph{Decision making}
\subsection{LACR 2: Orientation}\label{sec:orient}
Starting at the skeleton output by LACR 1, we continue to determine the direction of each edge in the skeleton.
In LACR 2, we simply utilize direct query to LLM for the orientation task due to LLMs' high performance on causal orientation tasks \cite{kiciman2023causal}.
For each pair of adjacent variables in the skeleton, we use a two-step prompt strategy:\\
{\bf Background reminder}.
Similar to LACR 1, we provide the main variables and the domain information of the task, and ask the LLM to clarify the variables' meanings, as well as their interaction.\\
{\bf Orienting}. With the above clarification, we ask the LLM to thoroughly understand the given $\KB$ and a {\sc causal direction context},
that specifies that if variable $A$ is the cause of variable $B$, then, the change of $A$'s value causes a change of $B$'s value, but not vice versa.
Then, we ask the LLM to give its decision based on all of the above information.

%\paragraph{Prompt strategy}
\section{Experiments}\label{sec:experiment}
In this section, we first introduce the ground truth datasets and how we collect three research literature pools. Then we introduce the settings of our solution and baselines. Finally, we evaluate the pruning and orienting results, respectively.

%In this section, we implement our LACR in the CD task for the well established SACHS data set \cite{sachs2005causal}, compare the results of several configurations, and compare our results with a statistical method called FASK \cite{sanchez2018causal,ramsey2018fask}.

\subsection{Experiment Data}
\noindent\textbf{Validation datasets.}
We validate our method on four datasets (namely, ASIA, SACHS, and CORONARY). %three of which have small-sized graphs and one has a medium-sized graph.
All datasets have reported causal graphs (see Appendix~\ref{sec:causal_graph}) based on real-world data. It is worth noting that, we only limit the selection of validation datasets to real-world datasets because LACR uses a realistic knowledge base.
%Given that our method fully depends on real-world knowledge and research, we select these datasets because their ground-truth graphs are constructed from practical data.

\noindent {\bf ASIA \cite{lauritzen1988local}.}
The ASIA dataset has 8 nodes (from domains of medical, biology, and social science) and 8 edges, revealing the potential reasons and symptoms of lung diseases.

% The ASIA dataset has a small-sized directed graph, which reveals the potential reasons and symptoms of lung cancer and tuberculosis.
% The ground truth graph has 8 nodes and 8 edges.
% The factors are mainly from the domains of medicine, biology, and social science.
% Note that there is one node, namely the ``Either Lung Cancer Or Tuberculosis'' node, acting as a logical link because the symptoms of ``Positive X-ray'' and ``Dyspnoea'' do not discriminate the two lung diseases.

\noindent {\bf SACHS \cite{sachs2005causal}.}
The SACHS dataset has 11 nodes (from the medical and biological domains) and 16 edges. It uncovers the interaction among proteins related to several human diseases.
% The SACHS dataset has a small-sized directed graph that studies the interaction between proteins that potentially cause human diseases.
% We use the ground truth graph given by the biologist domain, which is also used to refine the graph obtained by a Bayesian learning method in \cite{sachs2005causal}.
% The graph has 11 nodes and 16 edges,
% %(\red{recheck! \yd{I counted the edges again. The result is 14, while 1 of the edge is bidirectional.}}, 
% and all nodes are the abbreviations of protein names.
%\yz{I counted the edges in the Biologist's graph and here are the edges [['PIP3', 'PIP2'], ['PIP3', 'PLC'], ['PIP3', 'AKT'], ['PLC', 'PIP2'], ['PLC', 'PKC'], ['PIP2', 'PKC'], ['PKC', 'JNK'], ['PKC', 'P38'], ['PKC', 'RAF'], ['PKA', 'AKT'], ['PKA', 'JNK'], ['PKA', 'P38'], ['PKA', 'RAF'], ['RAF', 'MEK'], ['MEK', 'ERK'], ['PKA', 'ERK']]. Could you double check the list @Yidong? Thanks!}

\noindent {\bf CORONARY \cite{reinis1981prognostic}.}
The CORONARY dataset has 6 nodes (from the medical and biological domains) and 9 {\em undirected} edges, revealing the causal relationship among several potential reasons of coronary heart disease.
We only use it to validate LACR 1 because the edges are undirected.

\subsection{Experimental Settings}
We use GPT-4o in the following experiments.
%with the temperature set to $0$. \yd{The reviewers may question this choice. One potential solution is to do a validation study on a single dataset.}
%\yz{Or maybe we don't mention the setting? I don't remember I saw many such details in other LLM-related papers.}

\noindent {\bf Research document pool construction.}
In our experiment, we automatically build the pre-retrieved document set for each variable pair ({\bf Initialization} in Algorithm~\ref{algo:LACR 1}) in two steps:

\noindent (1) Relevant paper search: 
We search 20 paper titles by querying ``name[$v_i$] and name[$v_j$]'' to the Google Scholar engine using the SerpApi \cite{SerpApi}, and rank the papers by Google Scholar's default relevance ranking.
%We use the SerpApi\footnote{https://serpapi.com/google-scholar-api} to retrieve 20 paper titles by the Google Scholar search engine.
%Specifically, for each pair of variables, we simply send a query ``name[$v_i$] and name[$v_j$]'' to the Google Scholar API and select the top 20 papers according to Google Scholar's default ranking criteria.

\noindent (2) Paper download: Based on the aforementioned ranked paper title list, we use the PubMed API\footnote{https://www.ncbi.nlm.nih.gov/home/develop/api/}  to download the papers.
For each paper title, we prioritize downloading the full document from the PubMed Central (PMC) database, and only download the abstract document from the PubMed  database if the full version is not available in PMC.
for each variable pair, we download up to $10$ documents from the top of the ranked title list (note that some papers are unavailable in PubMed).
%For each paper, we prioritize downloading the full paper from the PubMed Central (PMC) \cite{pmc} database, and only download the abstract from the PubMed \cite{pubmed} database if the full paper is not available in PMC.
%Note that there may be no return from either PMC or PubMed for some papers.
%Then, for each variable pair, we download up to $10$ documents (abstracts or full papers) that are most relevant according to Google Scholar's ranking.

{\noindent\bf Statistical causal discovery method.}
In the validation of LACR 1, additional to LLM, we also test the impact of injecting statistical estimation-based results into the decision-making phase.
That is, adding point $1$ (resp. $-1$) to score $S$ if the statistical estimation-based method determines $\hat{\zeta}_\KB (ij)=1$ (resp. $\hat{\zeta}_\KB (ij)=0$) in Algorithm~\ref{algo:LACR 1}, where $\KB$ is numerical data.
We use the Peter-Clark (PC)\cite{spirtes2001} algorithm as the statistical estimation-based method.
We import the data from the bnlearn package \cite{scutari2019package}.

{\noindent\bf Baseline methods.}
We survey recent LLM-based causal graph construction methods, and for each dataset, we select the baseline method with the best performance.
For each dataset, we present two types of baseline LLMs: baseline LLM1, which is a pure LLM-based method, and baseline LLM2, which is a hybrid method combining a statistical estimation-based and an LLM-based method.
We do not compare LACR to any baseline method on the CORONARY dataset as the dataset's absence in such methods' validation.
%We do not compare to any baseline LLM-powered method on the CORONARY dataset since no other LLM-powered method uses the CORONARY dataset in validation yet as far as our investigation.

{\noindent\bf Validation metrics.}
We measure LACR 1 and LACR 2 by different metrics.
For LACR 1, we show the the adjacency precision (AP), the adjacency recall (AR), the F1 score, and the Normalized Hamming Distance (NHD), as follows.
%We separately measure the performances of the two phases of our method, i.e., the edge existence verification phase and the orientation phase.
%In each phase, we compute the adjacency precision (AP), the adjacency recall (AR), the F1 score, and the Normalized Hamming Distance (NHD).
%To compute the AP and AR, we first count three indices of the constructed graph:

First, we count three attributes of each graph:
true positive ($\mathtt{TP}$): the number of edges that are successfully recovered, false positive ($\mathtt{FP}$): the number of edges that are recovered but different from the ground truth graph, and false negative ($\mathtt{FN}$): the number of edges that exist in the ground truth but not recovered in our constructed graph.
Then, we compute
AP: $\frac{\mathtt{TP}}{\mathtt{TP}+\mathtt{FP}}$,
AR: $\frac{\mathtt{TP}}{\mathtt{TP}+\mathtt{FN}}$,
F1: $\frac{2AP*AR}{AP+AR}$, and
NHD: $\frac{\mathtt{FP}+\mathtt{FN}}{n^2}$, where $n$ is the number of variables.
Intuitively, NHD is the number different edges between two graphs, normalized by $n^2$.
% \noindent $\bullet$ True positives ($\mathtt{TP}$): the number of edges that are successfully recovered.\\
% $\bullet$ False positives ($\mathtt{FP}$): the number of edges that are recovered but different from the ground truth graph.\\
% $\bullet$ False negatives ($\mathtt{FN}$): the number of edges that exist in the ground truth but not recovered in our constructed graph.

% Then, we compute the metrics as:
% AP: $\frac{\mathtt{TP}}{\mathtt{TP}+\mathtt{FP}}$,
% AR: $\frac{\mathtt{TP}}{\mathtt{TP}+\mathtt{FN}}$,
% F1: $\frac{2AP*AR}{AP+AR}$, and
% NHD: $\frac{\mathtt{FP}+\mathtt{FN}}{n^2}$, where $n$ is the number of variables.
% Intuitively, NHD is the number different edges between two graphs, normalized by $n^2$.

In the validation of LACR 2, we simply compute the True Edge Accuracy (TEA), i.e., the ratio of correctly oriented edges among all true positive edges in LACR 1's output skeleton.

\subsection{Evaluation}
\begin{table}[t]
\centering
\small
\begin{tabular}{|l|l|l|l|l|l|}
\hline
\multicolumn{1}{|l|}{} & \multicolumn{1}{l|}{Dataset} & \multicolumn{1}{l|}{AP}    & \multicolumn{1}{l|}{AR}    & \multicolumn{1}{l|}{F1}    & SHD \\ \hline
\multirow{5}{*}{\rotatebox{90}{ASIA}} 
& LACR 1 ($\BACK$)        & \textbf{1}     & \textbf{1}     & \textbf{1}     &  0   \\ \cline{2-6} 
& LACR 1 ($\DOC$)        & 0.571  & \textbf{1}     & 0.727  &  0.122   \\ \cline{2-6} 
& LACR 1 ($\PC$)     & \textbf{1}     & 0.75  & 0.857  & 0.041    \\ \cline{2-6} 
& Baseline LLM1 & \textbf{1}     & 0.88  & 0.93  &   0.016   \\ \cline{2-6}
& Baseline LLM2 & 0.8     & \textbf{1}  & 0.89  &   0.031   \\ \hline

\multirow{3}{*}{\rotatebox{90}{CORO}} 
& LACR 1 ($\BACK$)        & 0.625 & 0.625 & 0.625 &  0.167   \\ \cline{2-6} 
& LACR 1 ($\DOC$)        & 0.667  & 0.75  & 0.706  &  0.139   \\ \cline{2-6} 
& LACR 1 ($\PC$)     & \textbf{0.778} & \textbf{0.875} & \textbf{0.824} &  0.083   \\ \hline

\multirow{5}{*}{\rotatebox{90}{SACHS}} 
& LACR 1 ($\BACK$)        & \textbf{0.8}      & 0.5      & 0.615      &  0.083   \\ \cline{2-6} 
& LACR 1 ($\DOC$)        & 0.467      & \textbf{0.875}      & \textbf{0.609}      & 0.149    \\ \cline{2-6} 
& LACR 1 ($\PC$)     & 0.421      & 0.5      & 0.457      & 0.157    \\ \cline{2-6} 
& Baseline LLM1 & N/A  & N/A   & 0.31  &  0.63   \\ \cline{2-6}
& Baseline LLM2 & 0.59  & N/A   & 0.56  &  0.12   \\ \hline

% \multirow{4}{*}{\rotatebox{90}{CHILD}}
% & LACR 1 ($\BACK$)        & 0.514      & 0.72      & 0.6      &  0.06   \\ \cline{2-6} 
% & LACR 1 ($\DOC$)        & 0.306      & 0.44      & 0.361      &  0.098   \\ \cline{2-6} 
% & LACR 1 ($\PC$)     & 0.4      & 0.4      & 0.4      & 0.075    \\ \cline{2-6} 
% & Baseline LLM & 0.59      & 0.68      & 0.63      &   0.055   \\ \hline
\end{tabular}
\caption{Performances of our solution LACR 1 with different $\KB$. We test the performance across three datasets, and compare to baseline methods: ASIA: LLM1: \cite{jiralerspong2024efficient}, LLM2: \cite{jiralerspong2024efficient},
SACHS: LLM1: \cite{zhou2024causalbench}, LLM2: \cite{takayama2024integrating}.
%CHILD: LLM+statistical tests on 10000 data samples in \cite{jiralerspong2024efficient}.
}
\label{tab:LACR 1}
\end{table}
We now first present observations based on experimental results for three datasets, which contains Edge Existence Verification (Section~\ref{sec:exp:existence}) and Orientation (Section~\ref{sec:exp:orientation}), followed by a comprehensive analysis of the overall results (Section~\ref{sec:exp:result}).

\subsubsection{Observation on Edge Existence Verification}~\label{sec:exp:existence}
We present the performance of LACR 1 on causation existence verification with different knowledge bases $\KB$ in the section. The orienting performance will be shown in the next section. Table~\ref{tab:LACR 1} lists the performance of all compared methods, where $\BACK$ denotes only LLM's background knowledge, $\DOC$ denotes both LLM's background knowledge and the fixed number of documents, and $\PC$ denotes $\DOC$ plus the results output by the PC algorithm. We have the following observations:

%\subsubsection{LACR 1: Causation Existence Verification}\label{sec:exp:existence}
%As shown in Table~\ref{tab:LACR 1}, we list the performance of LACR 1 with different knowledge bases $\KB$, where $\BACK$ denotes only LLM's background knowledge, $\DOC$ denotes both LLM's background knowledge and the fixed number of documents, and $\PC$ denotes $\DOC$ plus the results output by the PC algorithm.

%It is worth observing that on the ASIA dataset and the SACHS dataset, LACR 1 outperforms the selected LLM-powered methods with specific $\KB$ settings.
%As follows, we illustrate the result for each dataset.

{\noindent\bf ASIA.}
We have three observations from the experimental results on the ASIA dataset. 
First, LACR 1 achieves the best performance when relying solely on $\BACK$. It successfully recovers the full skeleton and outperforms the high performance of the pure LLM method in \cite{jiralerspong2024efficient}. 
Second, adding retrieved documents into $\KB$ reduces performance (AP from $1$ to $0.57$, and F1 score from $1$ to $0.73$) according to the given ground truth in \cite{lauritzen1988local}. 
%As we will explain in Section~\ref{sec:new_truth}, this drop is due to the ground truth being outdated, as domain knowledge has advanced.
%
Third, by further aggregating the output of the PC algorithm, the F1 score increases from $0.73$ to $0.86$ compared to the ground truth in \cite{lauritzen1988local}.
%It is because the PC algorithm extracts the conditional associational information from the numerical data, where the joint probability distribution follows the Markov property of the ground truth causal graph, as stated in the bnlearn package.

%LACR 1 with $\KB$ of only LLM's background knowledge has the best performance, successfully recovering the full skeleton.
%It also outperforms the high performance of the pure LLM method in \cite{jiralerspong2024efficient}.
%However, by adding retrieved documents as the knowledge base, LACR 1's performance is reduced (AP from $1$ to $0.57$, and F1 score from $1$ to $0.73$), according to the given ground truth in \cite{lauritzen1988local}.
%We later reason in Section~\ref{sec:new_truth} that this performance dropping is due to the knowledge updating of the domain research, which can be strong evidence to update the ground truth.
%By further aggregating the output of the PC algorithm, the F1 score increases to $0.86$ from $0.73$ towards the ground truth of \cite{lauritzen1988local}.
%This is not surprising because the PC algorithm extracts the conditional associational information from the numerical data, of which the joint probability distribution follows the Markov property of the ground truth causal graph as stated in the data resource bnlearn package.

{\noindent\bf CORONARY (CORO).}
%We have one observation from the experimental results on the CORONARY dataset. 
The results differ notably from those based on the ASIA dataset, LACR 1 with only the LLM's background knowledge achieves the worst performance, with values of $0.625$ for all of AP, AR, and F1 scores.
By adding documents and the PC algorithm into $\KB$, all metrics increase, reaching $0.875$.
%This indicates that the ground truth for CORONARY is not as outdated compared to ASIA, allowing extra knowledge from documents or the PC algorithm to improve performance.
%Similarly, we show strong evidence from the domain research, showing that an update of the ground truth is necessary in Section~\ref{sec:new_truth}.

%Different from the trends on the ASIA dataset, the lowest performance of LACR is based on the $\KB$ of only LLM's backjground knowledge, with value $0.625$ for all of AP, AR, and F1 score.
%By adding $\KB$ of documents and the PC algorithm, all metrics increase, reaching $0.875$ for all AP, AR, and F1 score.
%This coincides with the intuition that the LLM performs better with more knowledge bases.
%However, as described in Section~\ref{sec:new_truth}, we also found strong evidence from the domain research, showing that update of the ground truth is necessary.

{\noindent\bf SACHS.}
We have three observations from the results on the SACHS dataset.
First, the best performance of LACR 1 is achieved using only the LLM's background knowledge, outperforming the baseline method (combined method of LLM and hybrid statistical methods of DirectLiNGAM).
Second, adding documents as the knowledge base slightly decreases the F1 score by $0.01$.
%After checking LLM's responses, we conclude two major reasons: 
%
%(1) it is hard for LLM to comprehend the highly professional descriptions in biological scientific papers that contain a vast amount of domain-specific terminology, which brings noise into the decision-making phase, leading to worse performance. 
%
%(2) the ground truth causal graph is out-of-date for the knowledge base from SOTA literature. 
%Another major reason of the worse performance might be the knowledge updating in the literature, suggesting an update on the ground truth causal graph.
%
Third, with the PC algorithm's output, the F1 score reduces to $0.46$, which is worse than the baseline method.
%This decline is attributed to the limited performance of the PC algorithm on the SACHS dataset.
% \paragraph{CHILD}
% TBA ...
%%%%%%%%%%%%%%%%%%%%% with CHILD %%%%%%%%%%%%%%%%%%%%%
% \subsubsection{LACR 2: Orienting}\label{sec:exp:orientation}
% \begin{table}[t]
% \centering
% \small
% \begin{tabular}{|l|l|l|l|}
% \hline
%              & ASIA & SACHS & CHILD \\ \hline
% LACR2 ($\BACK$)        &  1    &  1     & 0.61      \\ \hline
% LACR2 ($\DOC$)       &  1    &  1     & 0.91      \\ \hline
% \end{tabular}
% \caption{The TEA of LACR 2 on datasets of ASIA, SACHS, and CHILD.}
% \label{tab:lacr2}
% \end{table}
%%%%%%%%%%%%%%%%%%%%%%%%%%%%%%%%%%%%%%%%%%%%%%%%%%%%%%
\subsubsection{LACR 2: Orientation}\label{sec:exp:orientation}
\begin{table}[t]
\centering
\small
\begin{tabular}{|l|l|l|}
\hline
             & ASIA & SACHS  \\ \hline
LACR2 ($\BACK$)        &  1    &  1        \\ \hline
LACR2 ($\DOC$)       &  1    &  1      \\ \hline
LACR2 ($\PC$)       &  1    &  1      \\ \hline
\end{tabular}
\caption{The TEA of LACR 2 on datasets of ASIA, SACHS, based on LACR 1's output skeleton on $\KB$ of $\BACK$, $\DOC$, and $\PC$, respectively.}
\label{tab:lacr2}
\end{table}
The results of LACR 2, shown in Table~\ref{tab:lacr2}, show that TEA is $1$ (i.e., orienting all $\mathtt{TP}$ edges correctly) on both ASIA and SACHS, based on all $\KB$. 
We can observe that, upon successfully recovering causal edges by LACR 1, the orientation accuracy is high, reaching $1$ for all knowledge bases and all datasets.
It demonstrates the efficacy of the orientation prompt as well as LLM's capability for causal orientation reasoning.
We conjecture that the success of this task strongly depends on the rich evidence stored in the scientific literature, and the easy understandability of such evidence, compared to the extraction of associational relationship.

\subsubsection{Overall Results Analysis}\label{sec:exp:result}
By summarizing the overall performance of LACR, it is worth noticing the following points:

\noindent{\bf LACR's performance tends to monotonically increase by taking more $\KB$ with high quality and readability.}
Observing LACR's performance on ASIA and CORONARY, we notice that our methods generally perform better with high-quality and readability of the input documents and statistical results.
While we discuss the performance drop of LACR1 ($\DOC$) on the ASIA dataset later, the overall trend coincides with the Condorcet theorem \cite{Grofman1983ThirteenTI} in voting theory, which suggests that aggregating diverse, high-quality inputs leads to better outcomes.

\noindent{\bf LACR performs differently on tail and non-tail data.}
We observe that LACR performs better on ASIA and CORONARY datasets compared to the SACHS dataset. This is because the terms in ASIA and CORONARY are more common to LLMs during training. In contrast, SACHS mainly contains symbols with specific meanings in a specific area.
Despite feeding scientific documents to LACR on all datasets, the lack of prior knowledge of these symbols in the training phase limits LLMs' understanding of their meanings, resulting in hallucinations. This has been observed in RAG-based legal research tools~\cite{magesh2024hallucination}. % variable names and document contents of ASIA and CORONARY are much easier to comprehend.
%However, the contents on SACHS dataset are significantly more difficult, even for researchers from different domains.

%\paragraph{LACR is sensitive to the quality of $\KB$.}
%Adding a low-quality knowledge base can significantly reduce the performance of LACR, e.g., the performance dropping of LACR 1 ($\PC$) on the SACHS dataset.\yp{it is the same as the first point}

\noindent{\bf Updating on the current ground truth causal graphs is necessary.}
We found, through LACR's responses, strong evidence from domain research (see details in Appendix~\ref{sec:refined_dataset}) indicating that an update to the ground truth causal graphs is necessary. For example, on the ASIA dataset, the ground truth being outdated led to reduced performance when using additional documents. Similarly, for the CORONARY dataset, the improvement in performance with added documents and the PC algorithm suggests that the ground truth for CORONARY is more current compared to ASIA.

%\paragraph{Variable name matters.} \yp{this can be explained by "readability"}

\subsection{Refining the Ground Truth (ASIA and CORONARY)}\label{sec:new_truth}

The ground truth causal graph needs refinement because it is outdated and significantly differs from current SOTA domain knowledge. Additionally, we aim to determine if the LLM can identify new causal relationships based on SOTA literature. In this part, we present the observations and analyses on the refined ground truth causal graphs based on the refined ASIA and CORONARY datasets. 
We first show the evidences returned by the LACR supporting the refinement of the ground truths in Section~\ref{sec:refined_dataset}, and then we discuss how refined graph truth affects performance from three perspectives: causal inferring based on LLM'S background knowledge $\BACK$, external literature $\DOC$, and statistic data $\PC$ in Section~\ref{sec:refine_validation}.

\subsubsection{Evidences of Ground Truth Refinement}\label{sec:refined_dataset}
\paragraph{ASIA}\hfill

\noindent {\bf Smoking and Tuberculosis}
In the documents \cite{wang2009review, horne2012association, kim2022association, wang2018association, gupta2022prevalence, lindsay2014association, amere2018contribution, alavi2012association} fed into LLM as the $\KB$, strong evidence shows that Smoking and Tuberculosis are associated, and the association cannot be eliminated by controlling the other variables in the ASIA dataset.
This conflicts against the conditional associational relationship between these two variables in the ground truth causal graph (Appendix~\ref{sec:causal_graph}), since both of the only two paths have a collider, which indicates that Smoking and Tuberculosis are independent from each other.
Based on the scientific evidence returned by LACR, a causal link exists between the two factors.

\noindent {\bf Bronchitis and X-ray}
Documents based on LACR's response, \cite{jin2023tropheryma, ntiamoah2021recycling} show that an association exists between Bronchitis and Positive X-ray.
\cite{chen2020diagnosis} further develops a deep-learning-based method to detect bronchitis directly from X-ray reports for children with age from 1-17 years old.
The evidence shows an association between the two variables, and the association is not mediated by the variable ``Smoking'' as shown in the causal graph.
Therefore, we add a causal edge between Bronchitis and X-ray in the ground truth causal graph.

\paragraph{CORONARY}\hfill

\noindent {\bf Strenuous Mental Work and Family Anamnesis Of Coronary Heart Disease}
According to the ground truth causal graph skeleton \cite{reinis1981prognostic}, there is a direct causal relationship between variable Strenuous Mental Work and variable Family Anamnesis Of Coronary Heart Disease.
However, according to the evidence returned by our method and the intuitive description in \cite{reinis1981prognostic}, we observe that this causal linkage should be removed with high probability.
By \cite{edwards2000introduction} (p.26), this edge is not intuitively expected though it is recovered by the Bayesian learning method.
Additionally, non of LACR's responses suggests the direct association between the two variables, and moreover, it returns evidences showing the positive probability to remove the edge.
\cite{wright2007family} shows that people with Family Anamnesis Of Coronary Heart Disease are easier to react to mental stress from work by higher Systolic Blood Pressure.
\cite{hintsa2010family} shows that the association between psychosocial factors at work and coronary heart disease is largely independent from the Family Anamnesis Of Coronary Heart Disease.

% \paragraph{Physical work and pressure}

\noindent {\bf Systolic Blood Pressure and Family Anamnesis Of Coronary Heart Disease}
LACR also returns evidence \cite{barrett1984family} showing that Family Anamnesis of Coronary Heart Disease and Systolic Blood Pressure are associated even after the adjustment of several variables including Smoking.
We therefore also add this edge between the two variables.

\subsubsection{Validation On The Refined Ground Truth}\label{sec:refine_validation}
\noindent{\bf Background Knowledge $\BACK$.}
The performance relying solely on background knowledge $\BACK$ varies between the two datasets. In the ASIAN dataset, all results slightly drop down, whereas in the dataset, results improve. A possible reason is that the background knowledge $\BACK$ related to ASIA is outdated, while the knowledge $\BACK$ for CORONARY is more current. Consequently, when the ground truth is updated based on new domain-specific knowledge, the outcomes are different significantly between the datasets.

\noindent{\bf External Literature $\DOC$.}
Incorporating $\DOC$ significantly increases performance for both datasets, as the refind ground truth better aligns with SOTA research trends, demonstrating the importance of up-to-date and relevant domain knowledge in improving model accuracy.

\noindent{\bf Statistic Data $\PC$.}
Different from the results of incorporating $\DOC$, adding $\PC$'s results consistently worsens performance, highlighting the PC-based solution is using an outdated dataset compared to updated SOTA knowledge. As the ground truth evolves to reflect current research advancements, the relative performance of $\PC$-based results diminishes. These findings emphasize the importance of up-to-date domain-specific knowledge for accurate causal graph recovery.

%In this section, we present the performance analysis of LACR 1 on refined ground truth using different $\KB$ settings, compared to baseline LLM-powered methods, on ASIAN and CORONARY datasets.
%Results relying solely on$\BACK$ show varied outcomes: slight deterioration in ASIAN and improvement in CORONARY. One possible reason is outdated $\BACK$ for ASIA and current $\BACK$ for CORONARY.
%Incorporating $\DOC$ significantly increases performance for both datasets, as the refined ground truth better aligns with SOTA research trends. However, adding PC's results consistently worsens performance, highlighting the outdated PC dataset compared to updated SOTA knowledge. These findings emphasize the importance of up-to-date domain-specific knowledge for accurate causal graph recovery.

\begin{table}[t]
\centering
\small
\begin{tabular}{|l|l|l|l|l|l|}
\hline
\multicolumn{1}{|l|}{} & \multicolumn{1}{l|}{Dataset} & \multicolumn{1}{l|}{AP}   & \multicolumn{1}{l|}{AR}    & \multicolumn{1}{l|}{F1}    & SHD \\ \hline

\multirow{3}{*}{\rotatebox{90}{ASIA}}
& LACR 1 ($\BACK$)        & 1     & 0.8     & 0.889     & 0.041   \\ \cline{2-6} 
& LACR 1 ($\DOC$)        & 0.714  & 1     & 0.833  & 0.082   \\ \cline{2-6} 
& LACR 1 ($\PC$)     & 1     & 0.6  & 0.75  & 0.082    \\ \hline

\multirow{3}{*}{\rotatebox{90}{CORO}} 
& LACR 1 ($\BACK$)        & 0.75 & 0.75 & 0.75 & 0.111   \\ \cline{2-6} 
& LACR 1 ($\DOC$)        & 0.778  & 0.875  & 0.824  & 0.083    \\ \cline{2-6} 
& LACR 1 ($\PC$)     & 0.667 & 0.75 & 0.706 & 0.139    \\ \hline

\end{tabular}
\caption{Performances of LACR 1 on the refined ground truth with different $\KB$, comparing to baseline LLM-powered methods.}
\label{tab:LACR 1_refine}
\end{table}

\section{Conclusion}
In this paper, we proposed a novel LLM-based causal graph construction method called LACR which uses the constraint-based causal prompt strategy designed according to the constraint-based causal graph construction (CCGC) method.
Comparing to most existing LLM-based causal graph construction methods, that use the direct causal prompt to query LLMs to do highly complex causal reasoning, LACR mainly relies on LLMs to do low-complexity associational reasoning, and follows the process of CCGC to determine the causal relationships.
For accurate associational reasoning, we utilize LLMs' RAG feature to extract statistical evidence with high relevance and quality from a large scientific corpus.
Lastly, we validate LACR's efficacy on several well-known datasets and show LACR's outstanding performance among LLM-based methods.
More importantly, LACR's responses show the conflict between the ground truths and SOTA domain research, which requests a refinement of the validation ground truths.

\newpage

\section*{Limitations}
We first address three technical limitations of the current version of LACR.
The first is the paper search accuracy.
The pre-retrieved document set needs high quality and relevance to provide relevant evidence.
Therefore, we conjecture that using refined queries and other search engines can enhance the performance.
The second limitation is LLMs' understanding on highly professional documents.
Through our experiments, we found that LLMs' poor comprehension capability on specific domains, e.g., the SACHS dataset, limits LACR's performance.
An optional solution is to fine-tune LLMs to better understand such documents.
The third is the complexity of LACR.
The method needs to query each variable pair ($O(n^2)$), and for each variable pair, multiple documents need to be queried.

We then address other practical limitations.
What comes first is the need of up-to-data practical validation datasets and causal graphs in causal discovery community.
Many validation datasets are synthesized, which are not usable in such practical knowledge-based methods.
The second practical limitation is the access of scientific papers.
In our experiment, we focus on biomedical datasets for the accessibility of research papers in PubMed, however, the full contexts of most of the papers are not open accessible.
It would open the possibility of overall better understanding of causal relationships if full documents are accessible in more research domains and broader scientific databases.

% \bibliography{anthology,ACL'24/causal}
\bibliography{causal}

\clearpage

\appendix
\section{Appendix}
%\subsection{Example Appendix}
\label{sec:appendix}

\subsection{Examples}\label{sec:ex}

As follows, we first show an example of statistical estimation-based methods' vulnerability to a type of data bias, the so-called selection bias \cite{Bareinboim2014}.
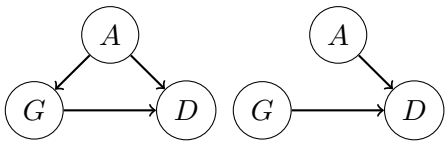
\begin{figure}[b]
\centering
\begin{tikzpicture}[->]
\node[circle,draw] (a) at (0,0) {$G$};
\node[circle,draw] (b) at (2,0) {$D$};
\node[circle,draw] (c) at (1,1) {$A$};
\node[circle,draw] (x) at (3,0) {$G$};
\node[circle,draw] (y) at (4,1) {$A$};
\node[circle,draw] (z) at (5,0) {$D$};

\path[thick] (c) edge (a);
\path[thick] (c) edge (b);
\path[thick] (a) edge (b);
\path[thick] (y) edge (z);
\path[thick] (x) edge (z);

\end{tikzpicture}
\caption{Causal graphs in Example~\ref{ex:bias}: left-the truth causal graph; right-recovered causal graph by the biased data.}
\label{fig:ex:bias}
\end{figure}
\begin{example}\label{ex:bias}
Consider that we would like to investigate the causal relationship of three variables: $A$ (human age), $G$ (human gender), and $D$ (some disease).
Assume that the true causal graph is the left figure in Figure~\ref{fig:ex:bias}.

Generally speaking, human age and gender are associated because female has a longer average lifespan.
Assuming that this association is only significant for $A\ge 60$.
However, if each point in a dataset has age under $60$, we cannot observe significant difference between the population of male and female.
Then, we would recover the causal graph as the right figure in Figure~\ref{fig:ex:bias}.
\end{example}
The second example shows the processing of a well-known constraint-based causal graph discovery algorithm called PC algorithm.
\begin{example}\label{ex:pc}
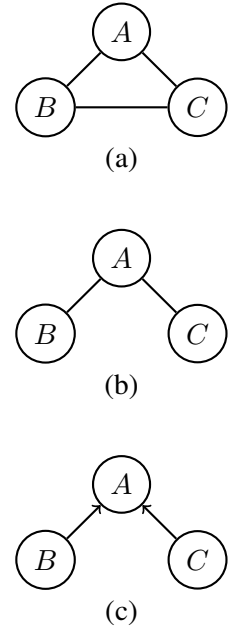
\begin{figure}
\centering
\begin{tikzpicture}[thick]
\node[circle, draw] (a1) at (5, 10) {$A$};
\node[circle, draw] (b1) at (4, 9) {$B$};
\node[circle, draw] (c1) at (6, 9) {$C$};
\path[draw] (a1) edge  (b1);
\path[draw] (a1) edge  (c1);
\path[draw] (b1) edge  (c1);
\node (d1) at (5, 8.3) {(a)};

\node[circle, draw] (a2) at (5, 7) {$A$};
\node[circle, draw] (b2) at (4, 6) {$B$};
\node[circle, draw] (c2) at (6, 6) {$C$};
\path[draw] (a2) edge  (b2);
\path[draw] (a2) edge  (c2);
\node (d2) at (5, 5.3) {(b)};

\node[circle, draw] (a3) at (5, 4) {$A$};
\node[circle, draw] (b3) at (4, 3) {$B$};
\node[circle, draw] (c3) at (6, 3) {$C$};
\path[->] (b3) edge  (a3);
\path[[->] (c3) edge  (a3);
\node (d3) at (5, 2.3) {(c)};

\end{tikzpicture}
\caption{PC algorithm's process.}
\label{fig:ex:pc}
\end{figure}
Consider a causal discovery task for three variables $A$, $B$, and $C$, and two different joint probability distributions $P^1$ and $P^2$.
We start with a complete undirected graph Figure (a) \ref{fig:ex:pc}.

Then, by $P^1$, we conduct the zero-order independence tests and obtain:
$\hat{\alpha}(AB)=1$, $\hat{\alpha}(AC)=1$, and $\hat{\alpha}(BC)=0$.
Then, we keep edges $(A,B)$ and $(A,C)$, and remove $(B,C)$, and obtain Figure (b) \ref{fig:ex:pc}, since $B$ and $C$ are not a cause of each other, otherwise they must be associated.
Based on the zero-order tests, we can already determine the causal graph as Figure (c) \ref{fig:ex:pc}, as $A$ must be a collider since $B$ and $C$ are d-separated by $\emptyset$.

On the other hand, if we consider $P^2$, we first have zero-order tests showing all pairs are associated, and we cannot remove any edge in Figure (a) \ref{fig:ex:pc}.
We then conduct first-order tests, and obtain: $\hat{\alpha}(AB\mid C)=1$, $\hat{\alpha}(AC\mid B)=1$, and $\hat{\alpha}(BC\mid A)=0$.
Therefore, we can remove the edge $(B,C)$ from Figure (a) \ref{fig:ex:pc}, and obatin Figure (b) \ref{fig:ex:pc}.
However, we cannot determine the directions of the edges because all directions of $A\rightarrow B \rightarrow C$, $A\leftarrow B \leftarrow C$, $A\leftarrow B \rightarrow C$ indicate the conditional independences consistent with $P^2$.
\end{example}

\newpage
\subsection{Additional Experiment Details}\label{sec:causal_graph}

The ground truth causal graphs of all datasets in Section~\ref{sec:experiment}.

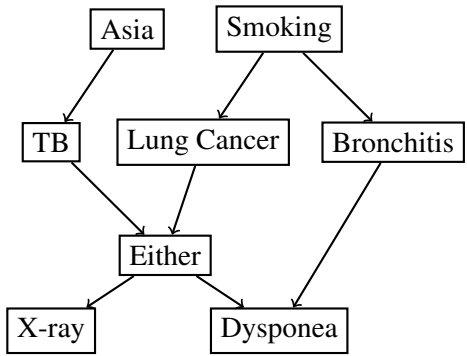
\begin{figure}[h]
\centering
\begin{tikzpicture}[thick, ->]
\node[rectangle,draw] (a) at (0,4) {Asia};
\node[rectangle,draw] (t) at (-1,2.5) {TB};
\node[rectangle,draw] (x) at (-1,0) {X-ray};
\node[rectangle,draw] (s) at (2,4) {Smoking};
\node[rectangle,draw] (l) at (1,2.5) {Lung Cancer};
\node[rectangle,draw] (e) at (0.5,1) {Either};
\node[rectangle,draw] (b) at (3.5,2.5) {Bronchitis};
\node[rectangle,draw] (d) at (2,0) {Dysponea};

\path (a) edge (t);
\path (t) edge (e);
\path (s) edge (l);
\path (s) edge (b);
\path (l) edge (e);
\path (b) edge (d);
\path (e) edge (x);
\path (e) edge (d);
\end{tikzpicture}
\caption{Ground truth causal graph of ASIA in \cite{lauritzen1988local}.}
\label{fig:asia_ori}
\end{figure}

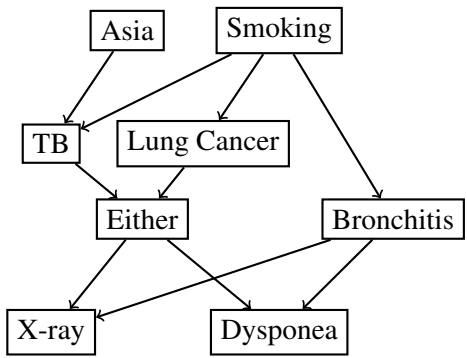
\begin{figure}[h]
\centering
\begin{tikzpicture}[thick, ->]
\node[rectangle,draw] (a) at (0,4) {Asia};
\node[rectangle,draw] (t) at (-1,2.5) {TB};
\node[rectangle,draw] (x) at (-1,0) {X-ray};
\node[rectangle,draw] (s) at (2,4) {Smoking};
\node[rectangle,draw] (l) at (1,2.5) {Lung Cancer};
\node[rectangle,draw] (e) at (0.2,1.5) {Either};
\node[rectangle,draw] (b) at (3.5,1.5) {Bronchitis};
\node[rectangle,draw] (d) at (2,0) {Dysponea};

\path (a) edge (t);
\path (t) edge (e);
\path (s) edge (l);
\path (s) edge (b);
\path (l) edge (e);
\path (b) edge (d);
\path (e) edge (x);
\path (e) edge (d);
\path (s) edge (t);
\path (b) edge (x);
\end{tikzpicture}
\caption{Refined ground truth causal graph of ASIA by LACR.}
\label{fig:asia_refine}
\end{figure}

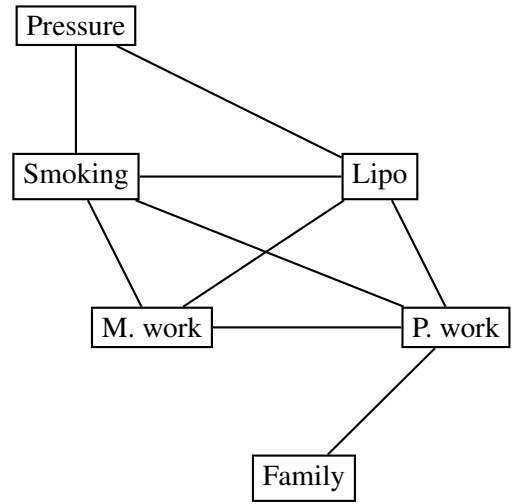
\begin{figure}[h]
\centering
\begin{tikzpicture}[thick, draw]
\node[rectangle,draw] (p) at (0,6) {Pressure};
\node[rectangle,draw] (s) at (0,4) {Smoking};
\node[rectangle,draw] (l) at (4,4) {Lipo};
\node[rectangle,draw] (m) at (1,2) {M. work};
\node[rectangle,draw] (pw) at (5,2) {P. work};
\node[rectangle,draw] (f) at (3,0) {Family};

\path (p) edge (s);
\path (p) edge (l);
\path (s) edge (m);
\path (s) edge (pw);
\path (s) edge (l);
\path (m) edge (pw);
\path (l) edge (pw);
\path (m) edge (l);
\path (pw) edge (f);

\end{tikzpicture}
\caption{Original ground truth causal graph of CORONARY in \cite{reinis1981prognostic}.}
\label{fig:coronary_ori}
\end{figure}

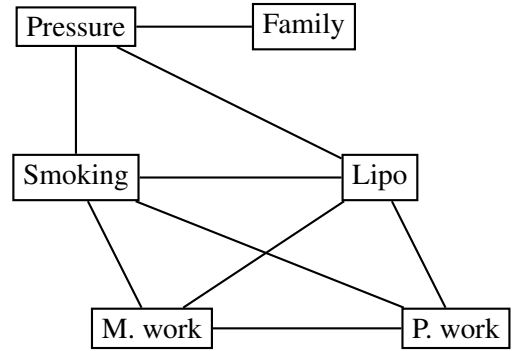
\begin{figure}[h]
\centering
\begin{tikzpicture}[thick, draw]
\node[rectangle,draw] (p) at (0,6) {Pressure};
\node[rectangle,draw] (s) at (0,4) {Smoking};
\node[rectangle,draw] (l) at (4,4) {Lipo};
\node[rectangle,draw] (m) at (1,2) {M. work};
\node[rectangle,draw] (pw) at (5,2) {P. work};
\node[rectangle,draw] (f) at (3,6) {Family};

\path (p) edge (s);
\path (p) edge (l);
\path (s) edge (m);
\path (s) edge (pw);
\path (s) edge (l);
\path (m) edge (pw);
\path (l) edge (pw);
\path (m) edge (l);
\path (p) edge (f);

\end{tikzpicture}
\caption{Refined ground truth causal graph of CORONARY by LACR.}
\label{fig:coronary_ori}
\end{figure}

% \begin{figure}[ht]
%   \centering
%   \includegraphics[width=1.00\columnwidth]{img/biologists_ground_truth.jpeg}
%   \caption{Original ground truth causal graph in \cite{sachs2005causal}.}
%   \label{fig:biologists_truth}
% \end{figure}

\begin{figure}[h]
\centering
\begin{tikzpicture}[thick, ->]
\node[rectangle,draw] (pkc) at (2,2) {PKC};
\node[rectangle,draw] (plc) at (0,4) {PLC$_\gamma$};
\node[rectangle,draw] (pip3) at (2,4) {PIP3};
\node[rectangle,draw] (akt) at (4,4) {AKT};
\node[rectangle,draw] (pip2) at (0,2) {PIP2};
\node[rectangle,draw] (pka) at (4,2) {PKA};
\node[rectangle,draw] (raf) at (6,4) {RAF};
\node[rectangle,draw] (mek) at (6,2) {MEK};
\node[rectangle,draw] (erk) at (6,0) {ERK};
\node[rectangle,draw] (p38) at (4,0) {P38};
\node[rectangle,draw] (jnk) at (2,0) {JNK};

\path (plc) edge (pkc);
\path (pip2) edge (pkc);
\path (pip3) edge (plc);
\path (pip3) edge (pip2);
\path (pip2) edge (pip3);
\path (plc) edge (pip2);
\path (pip3) edge (akt);
\path (pka) edge (akt);
\path (pka) edge (raf);
\path (pkc) edge (raf);
\path (raf) edge (mek);
\path (mek) edge (erk);
\path (pka) edge (erk);
\path (pka) edge (p38);
\path (pkc) edge (p38);
\path (pkc) edge (jnk);
\path (pka) edge (jnk);

\end{tikzpicture}
\caption{Biological ground truth causal graph in \cite{sachs2005causal}.}
\label{fig:biologists_truth}
\end{figure}
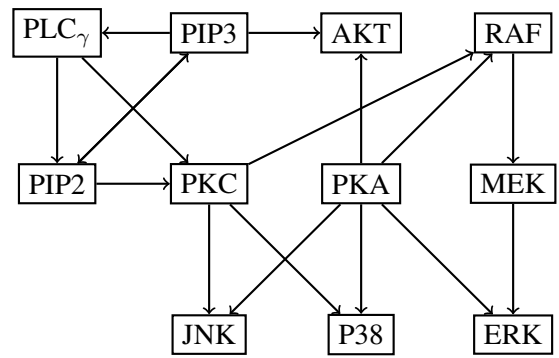

\clearpage

\subsection{Prompts}\label{sec:prompt}
% Original prompts are described in Section~\ref{sec:query}.
% Section~\ref{sec:prompbg} is the prompt to query LLM's background knowledge.
% Section~\ref{sec:promptdoc} is the prompt to use LLM to extract associational relationships from retrieved chunks.
% Then, in Section~\ref{sec:promptorbg} and Section~\ref{sec:promptordoc}, we give the prompts for the RAG-based orientation (Section~\ref{sec:orient}) from LLM's background knowledge and retrieved chunks.
\subsubsection{Association Context}
\scriptsize
\begin{lstlisting}[frame=single, breaklines, keepspaces, breakatwhitespace=false, breakindent=0pt]
The association relationship between two factors A and B can be associated or independent, and this association relationship can be clarified by the following principles:

1. If A and B are statistically associted or correlated, they are associated, otherwise they are independent.
2. The association relationship can be strongly clarified if there is statistical evidence supporting it.
3. If there is no obvious statistical evidence supporting the association relationship between A and B, it can also be clarified if there is any evidence showing that A and B are likely to be associated or independent statistically.
4. If there is no evidence to clarify the association relationship between A and B, then it is unknown.
\end{lstlisting}

\subsubsection{Association Type Context}
\begin{lstlisting}[frame=single, breaklines, keepspaces, breakatwhitespace=false, breakindent=0pt]
If two factors A and B are associated, they may be directly associated or indirectly associated with respect to a set of Given Third Factors, and it can be clarified by the following principle:

1. The first principle is to try to find statistical evidence from the given knowledge to clarify the following association types. If you cannot find statistical evidence, at lease find evidence that is likely to be able to statistically clarify the association type between A and B. If no obvious evidence can be found, the association type is unknown.
2. If the evidence shows that any factors from the Given Third Factors mediate the association between A and B, then A and B are indirectly associated via these factors.
3. If the evidence shows that by controlling any factors from the Given Third Factors, A and B are not associated any more, then A and B are associated indirectly.
4. If the evidence shows that A and B are still associated even if we control any of the given third factors, then A and B are directly associated.
5. If you think A and B are indirectly associated via any of the given third factors, it must be true that: (1) A and the third factors are directly associated; (2) B and the third factors are directly associated.
\end{lstlisting}

\subsubsection{Association Background Reminder}
\begin{lstlisting}[frame=single, breaklines, keepspaces, breakatwhitespace=false, breakindent=0pt]
As a scientific researcher in the domains of {domain}, you need to clarify the statistical relationship between some pairs of factors. You first need to get clear of the meanings of the factors in {factors}, which are from your domains, and clarify the interaction between each pair of those factors.
\end{lstlisting}

\subsubsection{LLM Association Query (with documents)}
\begin{lstlisting}[frame=single, breaklines, keepspaces, breakatwhitespace=false, breakindent=0pt]
Your task is to thoroughly read the given 'Document'. Then, based on the knowledge from the given 'Document', try to find statistical evidence to clarify the association relationship between the pair of 'Main factors' according to the 'Association Context' (delimited by double dollar signs).
Consider the given document and the association context. Answer the 'Association Question', write your thoughts, and give the reference in the given document. Respond according to the first expected format (delimited by double backticks).

Document:
{document}

Main factors:
{factorA} and {factorB}

Association Context:
$$
{association_context}
$$

Association Question:
Are {factorA} and {factorB} associated?

First Expected Response Format:
``
Document Identifier: XXX

Thoughts:
[Write your thoughts on the question]

Answer:
(A) Associated
(B) Independent
(C) Unknown

Reference:
[Skip this if you chose option C above. Otherwise, provide a supporting sentence from the document for your choice]
``
\end{lstlisting}

\subsubsection{LLM Association Type Query (with documents)}
\begin{lstlisting}[frame=single, breaklines, keepspaces, breakatwhitespace=false, breakindent=0pt]
Read and understand the Association Type Context. Consider carefully the role of any of the third factors appearing according to the Association Type Context. Then, based on your thoughts so far, answer the 'Association Type Question' with the 'Given Third Factors', write your thoughts, and give your reference in the given document. Respond according to the expected format (delimited by triple backticks)

Association Type Context:
$$$
{association_type_context}
$$$

Given Third Factors:
{factors} except for {factorA} and {factorB}

Association Type Question: Are {factorA} and {factorB} directly associated or indirectly associated?

Second Expected Response Format:
```
Thoughts:
[Write your thoughts on the question]

Answer:
(D) Directly Associated
(E) Indirectly Associated
(C) Unknown

Reference:
[Skip this if you chose option C above. Otherwise, provide a supporting sentence from the document for your choice]

Intermediary Factors:
[Skip this if you did not choose D or C above. Otherwise list all factors involved in this indirect association relationship, each separated by a comma]
```
\end{lstlisting}

\subsubsection{LLM Association Query (with background knowledge)}
\begin{lstlisting}[frame=single, breaklines, keepspaces, breakatwhitespace=false, breakindent=0pt]
Your task is to thoroughly use the knowledge in your training data to solve a task. Your task is: based on your background knowledge, try to find statistical evidence to clarify the association relationship between the pair of 'Main factors' according to the 'Association Context' (delimited by double dollar signs).
Consider your background knowledge and the association context. Answer the 'Association Question', and write your thoughts. Respond according to the 'First Expected Format' (delimited by double backticks).

Main factors:
{factorA} and {factorB}

Association Context:
$$
{association_context}
$$

Association Question:
Are {factorA} and {factorB} associated?

First Expected Response Format:
``
Thoughts:
[Write your thoughts on the question]

Answer:
(A) Associated
(B) Independent
(C) Unknown
``
\end{lstlisting}

\subsubsection{LLM Association Type Query (with background knowledge)}
\begin{lstlisting}[frame=single, breaklines, keepspaces, breakatwhitespace=false, breakindent=0pt]
Read and understand the 'Association Type Context'. Consider carefully the role of any of the third factors appearing according to the Association Type Context. Then, based on your thoughts so far, answer the 'Association Type Question' with the 'Given Third Factors', and write your thoughts. Respond according to the Second Expected Format (delimited by triple backticks)

Association Type Context:
$$$
{association_type_context}
$$$

Given Third Factors:
{factors} except for {factorA} and {factorB}

Association Type Question: Are {factorA} and {factorB} directly associated or indirectly associated?

Second Expected Response Format:
```
Thoughts:
[Write your thoughts on the question]

Answer:
(D) Directly Associated
(E) Indirectly Associated
(C) Unknown

Intermediary Factors:
[Skip this if you did not choose D or C above. Otherwise list all factors involved in this indirect association relationship, each separated by a comma]
```
\end{lstlisting}

\subsubsection{LLM Rethink Query}
\begin{lstlisting}[frame=single, breaklines, keepspaces, breakatwhitespace=false, breakindent=0pt]
If none of the Intermediary Factors you found is not in the Given Third Factor list, then, the association type between A and B is direct association.
Check your above response, and answer the Association Type Question again. Respond according to the Second Expected Format (delimited by triple backticks).

Given Third Factors:
{factors} except for {factorA} and {factorB}

Association Type Question: Are {factorA} and {factorB} directly associated or indirectly associated? 

Second Expected Response Format:
```
Thoughts:
[Write your thoughts on the question]

Answer:
(D) Directly Associated
(E) Indirectly Associated
(C) Unknown

Intermediary Factors:
[Skip this if you did not choose D or C above. Otherwise list all factors involved in this indirect association relationship, each separated by a comma]
```
\end{lstlisting}

\subsubsection{Causal Background Reminder}
\begin{lstlisting}[frame=single, breaklines, keepspaces, breakatwhitespace=false, breakindent=0pt]
As a scientific researcher in the domains of {domain}, you need to clarify the statistical relationship between some pairs of factors. You first need to get clear of the meanings of {factorA} and {factorB}, which are from your domains, and clarify the interaction between them.
\end{lstlisting}

\subsubsection{LLM Causal Direction Query (with background knowledge)}
\begin{lstlisting}[frame=single, breaklines, keepspaces, breakatwhitespace=false, breakindent=0pt]
Your task is to thoroughly use the knowledge in your training data to solve a task. Your task is: based on your background knowledge, try to find statistical evidence to clarify the direction of the causal relationship between the pair of 'Main factors' according to the 'Causal direction context' (delimited by double dollar signs).
Consider according to your background knowledge and the 'Causal direction context'. Answer the 'Causal direction question', and write your thoughts. Respond according to the 'Expected Format' (delimited by double backticks).

Main factors:
{factorA} and {factorB}

Causal direction context:
$$
{causal_direction_context}
$$

Causal direction question:
Is {factorA} the cause of {factorB}, or {factorB} the cause of {factorA}?

First Expected Response Format:
``
Thoughts:
[Write your thoughts on the question]

Answer:
(A) {factorA} is the cause of {factorB}
(B) {factorB} is the cause of {factorA}
(C) Unknown
```
\end{lstlisting}

\subsubsection{LLM Causal Direction Query (with documents)}
\begin{lstlisting}[frame=single, breaklines, keepspaces, breakatwhitespace=false, breakindent=0pt]
Your task is to thoroughly read the 'Given document' to solve a task. Your task is: based on the 'Given document', try to find statistical evidence to clarify the direction of the causal relationship between the pair of 'Main factors' according to the 'Causal direction context' (delimited by double dollar signs).
First thoroughly read and understand the Given document and the 'Causal direction context'. Then, Answer the 'Causal direction question', and write your thoughts. Respond according to the 'Expected Format' (delimited by double backticks).

Given document:
{document}

Main factors:
{factorA} and {factorB}

Causal direction context:
$$
{causal_direction_context}
$$

Causal direction question:
Is {factorA} the cause of {factorB}, or {factorB} the cause of {factorA}?

First Expected Response Format:
``
Thoughts:
[Write your thoughts on the question]

Answer:
(A) {factorA} is the cause of {factorB}
(B) {factorB} is the cause of {factorA}
(C) Unknown

Reference:
[Skip this if you chose option C above. Otherwise, provide a supporting sentence from the document for your choice]
\end{lstlisting}

\end{document}